%% file: iclr2027_conference.tex
\documentclass{article} %
\PassOptionsToPackage{nonamebreak}{natbib}
\usepackage{iclr2027_conference,times}

\input{math_commands.tex}

\usepackage[hidelinks]{hyperref}
\usepackage{url}
\usepackage{booktabs,tabularx,array,longtable,multirow,makecell}
\usepackage[table]{xcolor}
\usepackage{placeins,float,needspace}
\usepackage{algorithm,algpseudocode}
\usepackage{graphicx,amssymb}
\usepackage{subcaption}
\usepackage{wrapfig}
\definecolor{dhrow}{RGB}{242,239,250}
\definecolor{hpkDrugHarness}{HTML}{123A7A}
\definecolor{hpkBiomni}{HTML}{CB8C52}
\definecolor{hpkMarker}{gray}{0.45}
\DeclareRobustCommand{\hpkDrugHarnessKey}{\textcolor{hpkDrugHarness}{\rule{5pt}{5pt}}}
\DeclareRobustCommand{\hpkBiomniKey}{\textcolor{hpkBiomni}{\rule{5pt}{5pt}}}
\DeclareRobustCommand{\hpkMoleculeKey}{\textcolor{hpkMarker}{\ensuremath{\bullet}}}
\DeclareRobustCommand{\hpkMedianKey}{\textcolor{hpkMarker}{\raisebox{0.2pt}{\rotatebox[origin=c]{45}{\rule{4pt}{4pt}}}}}

\newcolumntype{Y}{>{\centering\arraybackslash}X}

\title{DrugHarness: An Evolvable Agent Harness for Computational Drug Discovery}

\author{%
\parbox{\dimexpr\textwidth-2\tabcolsep\relax}{%
\centering\normalfont
Yuxing~Chen$^{1,*}$,
Zhisheng~Zhang$^{2,*}$,
Cheng~Tang$^{1,3}$,
Wei~Li$^{1,4}$,
Minghui~Zhang$^{1,4}$,
Ziyan~Huang$^{1}$,
Shaocong~Wei$^{1}$,
Jiawei~Chen$^{1}$,
Junru~Hu$^{1,3}$,
Hongze~Zhu$^{5}$,
Jianghan~Shen$^{1,6}$,
Min~Cen$^{1}$,
Xinyuan~Zhu$^{7}$,
Yaxue~Zhao$^{4}$,
Jubo~Wang$^{2}$,
Yuqiang~Li$^{1}$,
Chun-Mei~Feng$^{8}$,
Zongyuan~Ge$^{9}$,
Rui~Su$^{10}$,
Jie~Xu$^{1}$,
Yirong~Chen$^{1}$,
Xi~Xu$^{2,\dagger}$,
Ming~Hu$^{1,\dagger}$
\endgraf\vspace{0.6em}
{\footnotesize
\mbox{$^{1}$Shanghai Artificial Intelligence Laboratory}\quad
\mbox{$^{2}$China Pharmaceutical University}\quad
\mbox{$^{3}$University of the Chinese Academy of Sciences}\quad
\mbox{$^{4}$Shanghai Jiao Tong University}\quad
\mbox{$^{5}$Westlake University}\quad
\mbox{$^{6}$Nanjing University}\quad
\mbox{$^{7}$Ant Group}\\
\mbox{$^{8}$Tongji University} \quad
\mbox{$^{9}$Monash University}\quad
\mbox{$^{10}$Shenzhen Loop Area Institute}
\endgraf\vspace{0.5em}
$*$ Equal contribution.
\qquad
$\dagger$ Corresponding authors.
\endgraf
}%
}%
}

\iclrfinalcopy %
\begin{document}
\raggedbottom

\maketitle

\input{sec/0_abstract}

\input{sec/1_intro}

\input{sec/2_related}
\input{sec/3_method}
\input{sec/4_experiment}

\input{sec/5_conclusion}

\bibliography{iclr2027_conference}
\bibliographystyle{iclr2027_conference}

\clearpage
\appendix
\input{sec/x_supp}

\input{sec/x_case_studies}

\end{document}

%% file: math_commands.tex
\usepackage{amsmath,amsfonts,bm}

\def\eqref#1{equation~\ref{#1}}

\def\1{\bm{1}}

\DeclareMathAlphabet{\mathsfit}{\encodingdefault}{\sfdefault}{m}{sl}
\SetMathAlphabet{\mathsfit}{bold}{\encodingdefault}{\sfdefault}{bx}{n}



%% file: sec/0_abstract.tex
\begin{abstract}

Computational drug discovery requires agents to compose scientific operations and adapt to changing evidence and constraints. Here we introduce DrugHarness, an evolvable agent harness coupling composable scientific execution with conditional skill induction. Its environment integrates over 80 scientific operations, biomedical resources and task-specific code, preserving associations among candidates, results and artifacts. From execution trajectories and outcome feedback, DrugHarness induces action hypotheses with observable applicability conditions, recommended actions and supporting evidence. An independent review agent evaluates candidate skills derived from these hypotheses through source verification and, when needed, targeted execution trials. Admitted skills guide subsequent tasks and are revised through feedback. Evaluation on MolBench and SMDD-Bench demonstrates capabilities in molecular screening, optimization and constrained design. Without evolution, DrugHarness exceeds reported MolClaw-CC results by 10.8 percentage points in affinity-comparison accuracy and 55\% in mean top-three active hits on MolBench. Retrospective evaluation on skill-development tasks shows that evolution increases Qwen’s affinity-comparison accuracy from 62.2\% to 81.1\% and optimization success from 79.5\% to 87.2\%. In a prospective HPK1 study, DrugHarness achieves a more favorable median docking score than Biomni among system-specific candidates. A synthesized compound achieves a cellular IC50 of 14.19 nM, a 2.64-fold potency improvement over the best input reference, with activity competitive with two selected compounds designed by experienced medicinal chemists. These results support conditional skill evolution for scientific agents and demonstrate a workflow connecting computational drug design with experimental validation.

\end{abstract}

%% file: sec/1_intro.tex
\section{Introduction}
\label{sec:introduction}
Drug discovery requires coordinating evidence and computation across molecular screening, optimization, and design. Experimental activity records, molecular representations, and property or binding predictors provide complementary information for candidate selection \citep{bindingdb2025,rdkit_software,admetai2024,boltz2_2025}. Turning these capabilities into a solution requires a sequence of decisions: which evidence to retrieve, which candidates to modify, which calculations to perform, and when a set of molecules satisfies the objective. Recent agent systems have begun to automate these multi-step tasks, making the organization of scientific actions and the reuse of effective strategies central questions for autonomous drug discovery \citep{averly2025liddia,wang2026cai}.

Progress spans three connected directions. \emph{Tool and environment integration} makes scientific resources accessible through specialized interfaces, code-based composition, and reusable infrastructure \citep{bran2023chemcrow,huang2025biomni,gao2025tooluniverse}. \emph{Workflow orchestration} organizes these operations through domain skills, specialist collaboration, and stateful processes that maintain intermediate results and guide subsequent actions \citep{zhang2026molclaw,gao2025pharmagents,cao2026mozi}. \emph{Experience and skill reuse} brings earlier solutions back into execution: prior designs inform new molecular proposals, historical tool trajectories guide lead optimization, and accumulated scientific experience supports later reasoning and tool selection \citep{gao2025pharmagents,li2026trace,tang2025chemagent,tang2026scitoolagentevo}. Together, these directions establish a basis for agents that both execute scientific workflows and accumulate guidance from their use.

Despite this progress, turning heterogeneous scientific experience into dependable action guidance presents three coupled challenges. \textbf{(1) Grounding a recommendation in the right evidence.} An activity measurement, a predicted property, an agent's interpretation, and a final task score answer different questions; their objects, endpoints, and observation times determine which action claims they support. \textbf{(2) Preserving applicability across tasks.} A useful source-task action may become inappropriate when the candidate state or remaining constraints change. For example, a set with adequate molecular properties but insufficient diversity may benefit from broader structural exploration rather than further optimization of similar candidates. \textbf{(3) Establishing executable and revisable guidance.} A plausible recommendation may rest on an unchecked scientific premise or an operation that behaves differently from its description. Its critical assumptions and execution path need active examination, and subsequent task feedback must inform revisions as new evidence becomes available. These challenges call for a learning process that connects scientific evidence, observable conditions, and active scientific review to the continuing use of a skill.

We introduce \textbf{DrugHarness}, a unified system for composable scientific execution and conditional skill learning. Its scientific environment provides over 80 scientific operations across ten capability categories, connecting tools, biomedical resources, and code while retaining associations among candidates, results, and artifacts. To ground reusable guidance, the learner compares these experiences and forms action hypotheses with explicit sources and applicability conditions. It translates each hypothesis into a skill whose conditions the executor assesses against the new task's visible state. An independent review agent examines evidence, applicability, and executability, using the same scientific tools and code environment for targeted checks when needed. Admitted skills become available to subsequent tasks, whose trajectories and native outcome feedback support further revision. Review records also allow rejected or deferred proposals to inform later changes. This design connects scientific judgment with the formation, use, and revision of action strategies.

We evaluate DrugHarness on two benchmarks and a prospective molecular design study. On MolBench, DrugHarness without evolution achieves 91.9\% affinity-comparison accuracy and 1.24 mean Hits@3, exceeding the reported MolClaw-CC results by 10.8 percentage points and 55\%, respectively. On SMDD-Bench, it satisfies all evaluation criteria in 24 of 40 lead-optimization tasks and supports structurally constrained design (Tables~\ref{tab:smdd} and~\ref{tab:molbench}). In retrospective evaluation on skill-development tasks, frozen learned skills increase Qwen's affinity-comparison accuracy from 62.2\% to 81.1\% and optimization success from 79.5\% to 87.2\% relative to historical runs without skills (Table~\ref{tab:evolution_summary}). In prospective HPK1 design, DrugHarness achieves a more favorable median docking score than Biomni among system-specific candidates (Figure~\ref{fig:hpk1_prospective}). A synthesized compound achieves a Jurkat p-SLP76 IC$_{50}$ of 14.19~nM, a 2.64-fold potency improvement over the best input reference, with activity competitive with two selected compounds independently designed by experienced medicinal chemists (Section~\ref{sec:case_studies}). These results demonstrate scientific execution capabilities, in-domain benefits from skill evolution, and experimentally validated molecular design.

Our main contributions are:
\begin{itemize}
    \item \textbf{Composable execution for molecular discovery.}
    DrugHarness integrates over 80 scientific operations, biomedical resources, and task-specific code, preserving associations among candidates, results, and artifacts to support adaptive scientific workflows. We will release the implementation, tool interfaces, and environment configurations.

    \item \textbf{Evidence-reviewed evolution of conditional skills.}
    DrugHarness induces action hypotheses with explicit applicability conditions and supporting evidence. An independent agent reviews candidate skills through source checks and targeted scientific trials when needed, while subsequent execution feedback refines their content and scope.

    \item \textbf{Evaluation from benchmarks to prospective validation.}
    We assess execution on two benchmarks, skill evolution through retrospective in-domain evaluation, and prospective HPK1 design through computational comparison with Biomni and experimental evaluation alongside reference and selected expert-designed compounds.
\end{itemize}

%% file: sec/2_related.tex
\section{Related Work}
\label{sec:related_work}
\paragraph{Scientific agents for drug discovery.}
Scientific agents organize capabilities through tool augmentation, code-based execution, and reusable environments, as illustrated by ChemCrow, Biomni, and ToolUniverse \citep{bran2023chemcrow,huang2025biomni,gao2025tooluniverse}. SciToolAgent further represents tool dependencies and compatibility in a knowledge graph for tool-chain planning \citep{ding2025scitoolagent}. Drug-discovery workflows use hierarchical domain skills, specialist collaboration, or stateful skill graphs to coordinate research stages \citep{zhang2026molclaw,gao2025pharmagents,cao2026mozi}; CAi and LIDDiA organize candidate evidence and feedback-driven molecular exploration \citep{wang2026cai,averly2025liddia}. Experience already participates in these workflows: PharmAgents summarizes earlier designs, and TRACE retrieves optimization trajectories to guide tool sequencing \citep{gao2025pharmagents,li2026trace}.

\paragraph{Experience-driven skill learning.}
Experience can be distilled into rules and examples, context-aware guidelines, or consolidated skills \citep{zhao2024expel,fu2024autoguide,trace2skill2026}. ACE maintains guidance incrementally, while SkillOpt and WikiSkill combine skill revision with execution-based selection and persistent learning state \citep{zhang2025agenticcontextengineeringevolving,skillopt2026,tang2026wikiskill}. Concurrent work on SkillCommit examines how replay-supported behavior governs skill consolidation and scope expansion \citep{he2026skillcommit}. Scientific applications learn different objects: ChemAgent maintains chemical subproblem solutions, DrugSAGE reuses experience in property-prediction model development, and SciToolAgent-Evo acquires scientific tool chains and situation-dependent selection guidance \citep{tang2025chemagent,drugsage2026,tang2026scitoolagentevo}. These studies provide complementary mechanisms for experience reuse across reasoning, modeling, and tool-use tasks.

%% file: sec/3_method.tex
\section{Method}\label{sec:method}

\subsection{System Overview and Scientific Execution Environment}\label{sec:system_overview}

\begin{figure}[t]
\centering
\includegraphics[width=\linewidth]{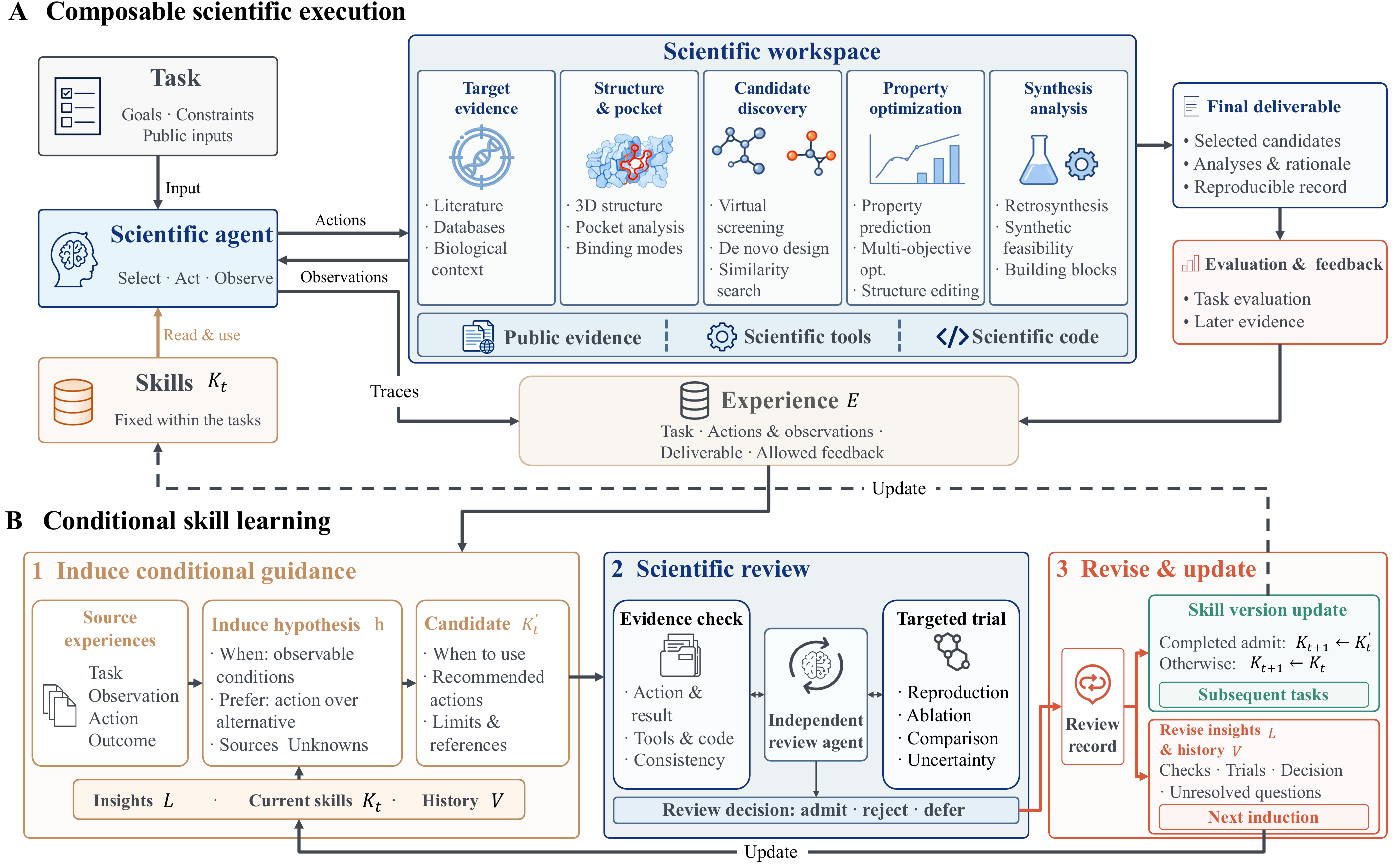}
\captionsetup{font=small}
\caption{\textbf{Overview of DrugHarness.} (A) The agent composes public evidence, scientific tools, and code into task-specific workflows around shared scientific objects. Traces and permitted outcome feedback form experiences. (B) The learner induces conditional action hypotheses and translates them into candidate skills. An independent reviewer checks their evidence and conducts targeted trials in the same workspace when needed. Admitted skills guide subsequent tasks, while review records update insights and history for further learning.}
\label{fig:system_overview}
\end{figure}

DrugHarness integrates composable scientific execution with conditional skill learning (Figure~\ref{fig:system_overview}). Task execution produces experiences for learning, and an independent review agent uses the same scientific environment to examine candidate skills. Admitted guidance supports subsequent tasks, whose feedback informs further revision.

DrugHarness provides over 80 agent-accessible scientific operations across ten capability categories, spanning evidence retrieval, target and structure preparation, molecular design, and computational evaluation (Table~\ref{tab:scientific_capabilities}). Biomedical resources and specialized tools such as BindingDB, RDKit, ADMET-AI, and Boltz-2 work alongside a scientific Python environment to support candidate assessment and design \citep{bindingdb2025,rdkit_software,admetai2024,boltz2_2025}.

In the shared scientific workspace, the agent selects and combines these operations according to the task's goals, constraints, and intermediate results. DrugHarness preserves associations among scientific objects, computational conditions, sources, and outputs across tool calls and task-specific code. As a result, evaluations from different tools remain linked to the same candidate, and returned structures or datasets can become inputs to further analysis. The runtime maintains the action loop as the agent uses this evidence to select further computations, revise candidates, and produce a final deliverable. Trajectories, deliverables, and permitted outcome feedback then provide material for skill learning.

\subsection{Inducing Conditional Scientific Strategies}\label{sec:strategy_induction}

Learning starts from recorded scientific decisions: the task being pursued, actions and observations during execution, the deliverable, and permitted outcome feedback. Each experience is
\begin{equation}
 e_i=(x_i,\tau_i,y_i,f_i),\qquad
 \tau_i=\bigl((a_{i,j},o_{i,j})\bigr)_{j=1}^{T_i},
 \label{eq:experience}
\end{equation}
where $x_i$ contains the task objective, constraints, and public inputs; $\tau_i$ records $T_i$ steps, with $a_{i,j}$ and $o_{i,j}$ denoting the action and visible observation at step $j$; $y_i$ is the deliverable; and $f_i$ is permitted outcome feedback. Distinguishing contemporaneous observations, the executor's interpretations, and subsequent evaluations lets the learner relate a decision to the evidence available at execution time and to feedback obtained afterward.

The learner identifies recurring decision problems across experiences and compares the relevant scientific objects, endpoints, methods, and evidence states. It examines decision-relevant conditions, such as which constraints have been met and whether the evidence needed for a decision has been obtained, rereading permitted sources when a specific question requires clarification.

Compatible conditions and consistent recommendations allow support to be combined; different conditions retain separate applicability scopes. For conflicting evidence under comparable conditions, the learner narrows or splits a recommendation when justified and retains unresolved disagreements. These comparisons connect the circumstances of an action to the evidence supporting a proposed change for a stated objective.

The result is a conditional action hypothesis specifying when an action should be preferred, what objective it serves, and the grounds for that choice:
\begin{equation}
 h=(g,c,a_0,a,R_h,U_h).
 \label{eq:hypothesis}
\end{equation}
Here $g$ is the goal and $c$ describes conditions observable by the executor; $a_0$ and $a$ denote the current or alternative action and the recommended action. Sources $R_h$ support or limit the hypothesis, and $U_h$ records unresolved information. For illustration, when individual molecules meet property requirements but set diversity is insufficient, a hypothesis may prioritize expanding structural sources over further optimization of similar candidates.

An LLM carries out this induction within a researcher-defined scope $\Omega$ of tasks, quality objectives, permitted materials, execution environment, and evaluation rules. At learning round $t$, it reads a batch $B_t\subseteq E$ from the experience collection $E$, together with accumulated insights $L_t$, current skills $K_t$, and revision history $V_t$. Under reading and generation allowance $\mathcal B_t$, the update is
\begin{equation}
 (\widetilde L_t,h_t)=\operatorname{Induce}(B_t,L_t,K_t,V_t;\Omega,\mathcal B_t).
 \label{eq:induce}
\end{equation}
The learner compares the proposed change with current skills and revision history to identify existing coverage and relevant prior findings. It returns revised insights $\widetilde L_t$ and a hypothesis $h_t$ for one proposed action change; when no reusable modification is warranted, it returns $h_t=\varnothing$ while updating insights. Algorithm~\ref{alg:induction} details this process.

\subsection{Candidate Skill Construction and Evidence Review}\label{sec:candidate_review}

The learner converts the action hypothesis into a skill update with one intended change in the executor's behavior. The revised guidance preserves the hypothesis's goal, applicability conditions, recommended action, and supporting evidence. Together with the remaining skills, it forms the complete candidate version $K'_t=\operatorname{Edit}(K_t,h_t)$. The skill's name and description support discovery; its body explains when and how to apply the recommendation, including relevant limits, while accompanying resources retain the scientific basis.

Each candidate states the decision situations it addresses and observable checks of its intended change. An independent review agent checks the supporting sources and scientific premises, asking whether they justify the recommendation under the stated conditions. It also examines whether the executor can assess those conditions from available information and carry out the instructions. The reviewer can retrieve original sources, inspect interfaces, and use the same scientific tools and code environment as the task executor. When a scientific or operational question requires computation, it conducts a targeted trial and retains the inputs, outputs, and limitations.

Review produces an admission decision and evidence for subsequent learning. A completed review returns
\begin{equation}
 \rho_t=\operatorname{Review}(K'_t,h_t,R_{h_t};\Omega)
 \in\{\mathrm{admit},\mathrm{reject},\mathrm{defer}\}.
 \label{eq:review}
\end{equation}
Admission makes the complete candidate version available for subsequent tasks. Conflicts with critical facts or constraints support rejection; insufficient evidence supports deferral. Completed source checks, trial findings, and unresolved questions remain available for later induction regardless of the decision.

\subsection{Task-time Use and Feedback-driven Revision}\label{sec:self_evolution}

For each candidate branch, review determines whether the current guidance $K_t$ is updated to the candidate version $K'_t$:
\begin{equation}
 K_{t+1}=\begin{cases}
 K'_t,&\text{review completed and }\rho_t=\mathrm{admit},\\
 K_t,&\text{otherwise}.
 \end{cases}
 \label{eq:deploy}
\end{equation}

Each task binds a specified skill version, whose contents remain fixed throughout that execution. The executor discovers relevant skills from their names and descriptions and reads their bodies and resources as needed. For a hypothesis $h$ and its condition $c$ expressed in the skill, applicability is assessed against the visible state $s_j$:
\begin{equation}
 \widehat\chi_j(h)=\operatorname{Assess}(c;s_j)
 \in\{\mathrm{supported},\mathrm{contradicted},\mathrm{unknown}\}.
 \label{eq:assess}
\end{equation}
The executor makes this semantic judgment from the available evidence. Supported conditions allow the recommendation to inform action selection; contradicted conditions direct the executor to other grounds for action. When a critical condition is unknown, the executor decides whether further information would help. These judgments inform which recommendations to use and how to adapt the actions to the current task.

Review and execution provide complementary inputs to the next learning round. Completed source checks, trial findings, and unresolved questions update insights $L$ and history $V$ whether or not a candidate is admitted. Subsequent task trajectories, deliverables, and permitted outcome feedback add new experiences to $E$ as they become available. Later induction draws on the updated experiences, insights, and history to reconsider action recommendations and their applicability conditions. Proposed changes then return to candidate construction and independent scientific review. Appendix~\ref{app:learning_algorithms} provides the learning loop and induction procedure.

%% file: sec/4_experiment.tex
\section{Experiments}
\label{sec:experiments}
We evaluate three aspects of DrugHarness: scientific execution without learned skills, execution with experience-derived skills, and prospective evaluation of designed molecules. Model and runtime comparisons support the execution analysis, while task trajectories clarify how learned guidance enters scientific decisions. Appendix~\ref{app:case_studies} presents four case studies of scientific execution, feedback use, and skill refinement.

\subsection{Experimental Setup}
\label{sec:experimental_setup}
\label{sec:benchmarks}
\paragraph{System configurations.}
\emph{DrugHarness (w/o Evol.)} uses the full scientific environment without cross-task learned skills; its main configuration is Codex~\citep{openai_codex} with GPT-5.6 Terra/xhigh. Model comparisons use Codex with Qwen3.7-Max, and runtime comparisons use Hermes~\citep{nousresearch_hermes_agent} with Terra/xhigh. Both comparisons cover all 190 MolBench tasks; runtime comparisons additionally cover 202 SMDD Types 1--5 tasks. We also evaluate the original SMDD harness with Terra/xhigh on the same SMDD task set. \emph{DrugHarness (Evol.)} uses admitted, frozen skills: Qwen3.8-Max/xhigh performs automatic refinement and independent review. DrugHarness uses a default execution budget of one hour per benchmark task and six hours for the HPK1 design task.

\paragraph{Evaluation tasks and metrics.}
We use the two benchmarks for complementary purposes. SMDD-Bench evaluates DrugHarness on diverse constrained molecular-design problems and serves as a broad test of scientific execution \citep{han2026smdd}. MolBench provides a fine-grained decomposition of molecular decision-making and operations, spanning property filtering, similarity filtering, affinity comparison, virtual screening, molecular editing, and optimization, together with task-specific quantitative feedback \citep{zhang2026molclaw}. We therefore use MolBench as the primary testbed for studying experience-driven evolution, where such fine-grained feedback enables systematic analysis of skill induction, use, and revision. SMDD-Bench uses task success rates; we evaluate a fixed 40-task lead-optimization (LO) subset spanning 25 targets and the complete remaining families. MolBench metrics include accuracy, mean active hits among the top three candidates (Hits@3), and positive-improvement success rate. Task breakdowns and metric definitions appear in Appendix~\ref{app:benchmark_details}. The prospective HPK1 task evaluates analog design under collaborator-provided references and constraints (Section~\ref{sec:case_studies}).

\subsection{Main Results}
\label{sec:main_results}

\input{tables/smdd_results}
\input{tables/molbench_combined}
\paragraph{Overall Performance.}
We use Codex with GPT-5.6 Terra as the main configuration to evaluate scientific execution without evolution. On SMDD-Bench, DrugHarness/Codex achieves 28.0\% success in interaction point discovery versus 0.0\% for the original SMDD harness, and 60.0\% versus 50.0\% on the same 40 lead-optimization tasks, with both systems using GPT-5.6 Terra (Table~\ref{tab:smdd}). On MolBench, affinity accuracy reaches 91.9\%, 10.8 percentage points above MolClaw-CC, and mean Hits@3 reaches 1.24 versus 0.80. Screening accuracy is 98.0\%, all 39 editing tasks are solved, and 37 of 39 optimization tasks yield positive improvement (94.9\%; Table~\ref{tab:molbench}). Together, these results establish support for candidate judgments, precise molecular operations, and designs with joint structural and property constraints.

\paragraph{Backbone and Runtime Comparisons.}
\label{sec:configuration_analysis}
\label{sec:model_comparison}
\label{sec:backend_comparison}
We vary the model within Codex to assess backbone dependence, and compare Hermes with Codex under Terra to assess runtime effects. With Qwen3.7-Max, DrugHarness attains 62.2\% affinity accuracy versus 54.1\% for both CAi and MolClaw, and 60.0\% Hit@3 versus 40.0\% and 52.0\%, respectively; screening (96.0\%) and editing (97.4\%) match CAi \citep{wang2026cai}. With Terra, both runtimes achieve 98.0\% screening and 100\% editing accuracy, while Codex leads Hermes in affinity accuracy (91.9\% versus 83.8\%) and mean Hits@3 (1.24 versus 0.92). Codex/Terra achieves the strongest overall performance among the DrugHarness configurations tested.

\subsection{Experience-driven Evolution}
\label{sec:evolution_analysis}
Beyond backbone and runtime choices, we examine whether execution experience can improve DrugHarness through reusable skills.

\input{tables/evolution_summary}

Domain-specific guidance raises screening accuracy to 100\%, affinity accuracy from 62.2\% to 81.1\%, and optimization success from 79.5\% to 87.2\%, while retaining 97.4\% editing accuracy (Table~\ref{tab:evolution_summary}). All 39 optimization tasks produce valid JSON submissions, compared with 34 in historical no-skill runs.

Learned guidance makes retrieved evidence actionable. In affinity task ms2-029, both skill versions led the executor to materialize the same collection of 933 records. The earlier guidance prompted text searches and structural reasoning; the revision prompted RDKit structure matching to associate candidates with activity records and distinguish $K_i$ from IC50. The decision changed from incorrect to correct. Across the domain, the learned skill corrects nine historical errors and retains twenty-one previously correct answers.

Optimization experience yields guidance for checking the actual deliverable. From formatting failures and their repairs, the learner forms instructions to read back and validate the saved answer before submission. The independent reviewer tests this procedure in a targeted execution trial before deployment; subsequent runs produce valid JSON on all 39 tasks, with positive property improvement on 34. The learned skills achieve 30/37 correct affinity comparisons and 34/39 successful optimizations, approaching the fixed researcher-written references of 33/37 and 35/39, respectively.

Virtual screening illustrates feedback-driven skill revision. The initial skill emphasized recovery of missing jobs, but runs exposed missing submissions and incomplete candidate coverage. DrugHarness used these trajectories and feedback to revise the skill. With the frozen revision, complete coverage of the original 60 candidates increased from 14/25 to 25/25 tasks. Revised runs achieved a top-three hit on 12 tasks and a mean reference rank of 26.01 versus the historical no-skill 29.05.

\subsection{Prospective Validation of Designed Molecules}
\label{sec:case_studies}
\label{sec:prospective_validation}
We design human HPK1 (MAP4K1) inhibitors from four experimentally characterized references, aiming to improve Jurkat p-SLP76 inhibition over the best reference (IC50: 37.47~nM). Designs preserve the shared core and specified double-bond geometry, modifying only terminal amide nitrogen substituents.\footnote{Candidate molecules are reported using coded identifiers; their chemical structures are withheld for confidentiality.} DrugHarness uses Codex/Qwen3.8-Max; Biomni uses its default configuration.

\begin{figure}[H]
\centering
\captionsetup{font=small,skip=3pt}
\begin{subfigure}[t]{0.38\linewidth}
\centering
\includegraphics[width=\linewidth]{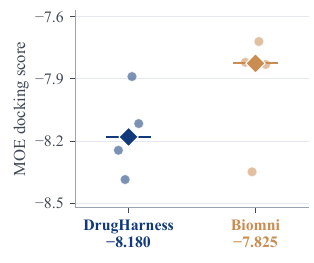}
\caption{MOE docking}
\label{fig:hpk1_docking}
\end{subfigure}\hfill
\begin{subfigure}[t]{0.60\linewidth}
\centering
\includegraphics[width=\linewidth]{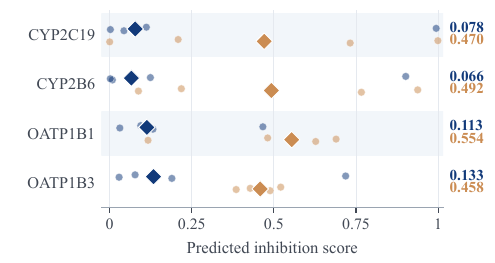}
\caption{Selected inhibition predictions}
\label{fig:hpk1_admet}
\end{subfigure}
\caption{\textbf{Computational evaluation of HPK1 candidates.} Four system-specific structures per group remain after excluding six shared structures from ten evaluated entries each. Colors: \hpkDrugHarnessKey\ DrugHarness; \hpkBiomniKey\ Biomni. Symbols: molecules (\hpkMoleculeKey), medians (\hpkMedianKey). Lower is favorable. The four inhibition endpoints were selected post hoc for lower DrugHarness medians; its median hERG inhibition was higher (0.949 versus 0.663) and QED lower (0.3645 versus 0.4085).}
\label{fig:hpk1_prospective}
\end{figure}
\vspace{-15pt}
\paragraph{In Silico Evaluation of Designed Molecules.}
Among the four system-specific candidates per evaluated set, DrugHarness achieves a more favorable median MOE docking score than Biomni ($-8.180$ versus $-7.825$; Figure~\ref{fig:hpk1_prospective}(\subref{fig:hpk1_docking})). It also has lower median inhibition predictions for CYP2C19 (0.078 versus 0.470), CYP2B6 (0.066 versus 0.492), OATP1B1 (0.113 versus 0.554), and OATP1B3 (0.133 versus 0.458; Figure~\ref{fig:hpk1_prospective}(\subref{fig:hpk1_admet})).

\vspace{-8pt}
\noindent
\begin{minipage}[b]{0.52\linewidth}
\paragraph{Experimental Validation of Designed Molecules.}
A DrugHarness-designed compound was synthesized and experimentally evaluated by our medicinal chemistry collaborators. It achieved a Jurkat p-SLP76 IC50 of 14.19~nM, lower than those of all four reference compounds (37.47--227.77~nM), together with 96.92\% HPK1 inhibition at 1~nM. On the same design task, the DrugHarness-designed compound showed competitive biochemical and cellular activity compared with two selected compounds independently designed by medicinal chemists with over five years of experience (Figure~\ref{fig:hpk1_chemist_comparison}).
\end{minipage}\hfill
\begin{minipage}[b]{0.46\linewidth}
\centering
\includegraphics[width=0.86\linewidth]{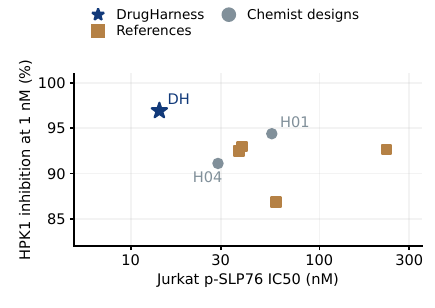}
\captionsetup{font=footnotesize}
\captionof{figure}{\textbf{Comparison with selected chemist designs.} Lower cellular IC50 and higher 1~nM HPK1 inhibition indicate stronger activity.}
\label{fig:hpk1_chemist_comparison}
\end{minipage}
\par

%% file: tables/smdd_results.tex
\begin{table}[!htbp]
\centering
\caption{\textbf{Results on SMDD-Bench.} Success rates (\%) on the same task set, including a fixed subset of 40 lead-optimization tasks.}
\label{tab:smdd}
\begingroup
\scriptsize
\setlength{\tabcolsep}{3pt}
\renewcommand{\arraystretch}{1.06}
\begin{tabularx}{0.85\linewidth}{llYYYYY}
\toprule
\multicolumn{1}{c}{Method / Runtime} &
\multicolumn{1}{c}{Model} &
\multicolumn{1}{c}{2D Pharm.$\uparrow$} &
\multicolumn{1}{c}{IP$\uparrow$} &
\multicolumn{1}{c}{SH$\uparrow$} &
\multicolumn{1}{c}{LO$\uparrow$} &
\multicolumn{1}{c}{FA$\uparrow$} \\
\midrule
\multirow{8}{*}{SMDD harness}
 & Claude Sonnet 4.6 & \textbf{28.0} & 0.0 & 3.8 & 47.5 & 0.0 \\
 & GPT-5.4 & 12.0 & 0.0 & 3.8 & 55.0 & \textbf{1.7} \\
 & Gemini 3.1 Pro & 20.0 & 4.0 & 0.0 & 47.5 & \textbf{1.7} \\
 & Kimi K2.5 Thinking & 12.0 & 0.0 & 1.9 & 32.5 & 0.0 \\
 & DeepSeek V3.2 & 8.0 & 0.0 & 3.8 & 22.5 & 0.0 \\
 & MiniMax M2.7 & 16.0 & 0.0 & 1.9 & 27.5 & 0.0 \\
 & Qwen3.5 397B A17B & 4.0 & 0.0 & 1.9 & 45.0 & 0.0 \\
 & GPT5.6-Terra$^{\dagger}$ & 16.0 & 0.0 & \textbf{9.6} & 50.0 & 0.0 \\
\midrule
\textbf{DrugHarness} / Hermes (w/o Evol.)
 & GPT5.6-Terra & 12.0 & 20.0 & 3.8 & 37.5 & 0.0 \\
\textbf{DrugHarness} / Codex (w/o Evol.)
 & GPT5.6-Terra & 12.0 & \textbf{28.0} & 7.7 & \textbf{60.0} & 0.0 \\
\bottomrule
\end{tabularx}
\par\smallskip
\begin{minipage}{0.85\linewidth}
\footnotesize
$^{\dagger}$ We ran the original SMDD harness with GPT-5.6 Terra. Other SMDD results are from \citet{han2026smdd}.
\end{minipage}
\endgroup
\end{table}

%% file: tables/molbench_combined.tex
\begin{table}[!t]
\centering
\caption{\textbf{Results on MolBench.} Baseline results are from \citet{zhang2026molclaw} and \citet{wang2026cai}. Metric definitions are provided in Appendix~\ref{app:molbench}.}
\label{tab:molbench}
\begingroup
\scriptsize
\setlength{\tabcolsep}{2pt}
\setlength{\aboverulesep}{1pt}
\setlength{\belowrulesep}{1pt}
\renewcommand{\arraystretch}{1.0}
\begin{tabularx}{0.95\linewidth}{@{}l|YYY|Y|YYY@{}}
\toprule
Method & \multicolumn{3}{c|}{MolClaw comparison group} & CAi & \multicolumn{3}{c}{\textbf{DrugHarness} (w/o Evol.)} \\
\cmidrule(lr){2-4}\cmidrule(lr){5-5}\cmidrule(l){6-8}
Model & \makecell[c]{Sonnet\\4.6} & \makecell[c]{Sonnet\\4.6} & \makecell[c]{Qwen3.7-\\Max} & \makecell[c]{Qwen3.7-\\Max} & \makecell[c]{Qwen3.7-\\Max} & \makecell[c]{GPT5.6-\\Terra} & \makecell[c]{GPT5.6-\\Terra} \\
Runtime & \makecell[c]{Claude\\Code$^\dagger$} & \makecell[c]{Claude\\Code} & N/R & CAi & Codex & Hermes & Codex \\
\midrule
\multicolumn{8}{c}{\textit{Screening}} \\
\midrule
Accuracy (\%)$\uparrow$ & 98.0 & 98.0 & 18.0 & 96.0 & 96.0 & 98.0 & \textbf{98.0} \\
Precision (\%)$\uparrow$ & \textemdash & \textemdash & 32.0 & 98.0 & 98.0 & 98.0 & \textbf{100.0} \\
Recall (\%)$\uparrow$ & \textemdash & \textemdash & 44.5 & 97.0 & 97.0 & 98.0 & \textbf{99.0} \\
F1 (\%)$\uparrow$ & \textemdash & \textemdash & 33.8 & 97.3 & 97.3 & 98.0 & \textbf{99.3} \\
Specificity (\%)$\uparrow$ & \textemdash & \textemdash & 44.0 & 98.0 & 100.0 & 98.0 & \textbf{100.0} \\
Validity (\%)$\uparrow$ & \textemdash & \textemdash & 78.0 & 100.0 & 98.0 & 100.0 & \textbf{100.0} \\
\midrule
\multicolumn{8}{c}{\textit{Affinity comparison}} \\
\midrule
Accuracy (\%)$\uparrow$ & 51.4 & 81.1 & 54.1 & 54.1 & 62.2 & 83.8 & \textbf{91.9} \\
Valid choice (\%)$\uparrow$ & \textemdash & \textemdash & 100.0 & 100.0 & 100.0 & 100.0 & \textbf{100.0} \\
\midrule
\multicolumn{8}{c}{\textit{Virtual screening}} \\
\midrule
Hit@3 (\%)$\uparrow$ & \textemdash & \textemdash & 52.0 & 40.0 & 60.0 & 48.0 & \textbf{60.0} \\
Mean Hits@3$\uparrow$ & 0.56 & 0.80 & 0.76 & 0.56 & 0.72 & 0.92 & \textbf{1.24} \\
Top-3 rate (\%)$\uparrow$ & \textemdash & \textemdash & 25.3 & 18.7 & 24.0 & 30.7 & \textbf{41.3} \\
Mean GT rank$\downarrow$ & \textemdash & \textemdash & 56.23 & 55.49 & 29.05 & 27.69 & \textbf{17.27} \\
Hit count$\uparrow$ & \textemdash & \textemdash & 13/25 & 10/25 & 15/25 & 12/25 & \textbf{15/25} \\
\midrule
\multicolumn{8}{c}{\textit{Editing}} \\
\midrule
Add (\%)$\uparrow$ & \textemdash & \textemdash & 70.0 & 100.0 & 100.0 & 100.0 & \textbf{100.0} \\
Delete (\%)$\uparrow$ & \textemdash & \textemdash & 88.9 & 100.0 & 100.0 & 100.0 & \textbf{100.0} \\
Substitute (\%)$\uparrow$ & \textemdash & \textemdash & 95.0 & 95.0 & 95.0 & 100.0 & \textbf{100.0} \\
Overall (\%)$\uparrow$ & 97.4 & 100.0 & 87.2 & 97.4 & 97.4 & 100.0 & \textbf{100.0} \\
Validity (\%)$\uparrow$ & \textemdash & \textemdash & 100.0 & 100.0 & 100.0 & 100.0 & \textbf{100.0} \\
\midrule
\multicolumn{8}{c}{\textit{Optimization}} \\
\midrule
Mean $\Delta$$\uparrow$ & 0.8660 & 1.7240 & 0.1352 & 0.2263 & 0.6054 & 0.4530 & 0.5988 \\
Success rate (\%)$\uparrow$ & \textemdash & \textbf{100.0} & 48.7 & 79.5 & 79.5 & 84.6 & 94.9 \\
Valid SMILES (\%)$\uparrow$ & \textemdash & \textemdash & 97.4 & 100.0 & 87.2 & \textbf{100.0} & 94.9 \\
Scaffold hard (\%)$\uparrow$ & \textemdash & \textemdash & \textbf{71.8} & 46.2 & 38.5 & 59.0 & 35.9 \\
Scaffold soft (\%)$\uparrow$ & \textemdash & \textemdash & \textbf{88.9} & 76.5 & 61.8 & 77.0 & 63.6 \\
\bottomrule
\end{tabularx}
\par\smallskip
\begin{minipage}{\linewidth}
\footnotesize N/R: Runtime not reported. $\dagger$: Claude Code without MolClaw skills.
\end{minipage}
\endgroup
\end{table}

%% file: tables/evolution_summary.tex
\begin{table}[htbp]
\centering
\caption{\textbf{Experience-driven evolution on MolBench.} Codex/Qwen3.7-Max with historical no-skill runs and retrospective in-domain evaluation using frozen domain-specific skills. Proportions are percentages. Complete metrics: Appendix~\ref{app:evolution_details}.}
\label{tab:evolution_summary}
\begingroup
\footnotesize
\setlength{\tabcolsep}{3pt}
\renewcommand{\arraystretch}{1.05}
\begin{tabular*}{0.85\linewidth}{@{\extracolsep{\fill}}lcccccccc@{}}
\toprule
Setting & Screening & Affinity & \multicolumn{3}{c}{Virtual screening} & Editing & \multicolumn{2}{c}{Optimization} \\
\cmidrule(lr){2-2}\cmidrule(lr){3-3}\cmidrule(lr){4-6}\cmidrule(lr){7-7}\cmidrule(l){8-9}
 & Acc.$\uparrow$ & Acc.$\uparrow$ & Hit@3$\uparrow$ & \makecell{Mean\\Hits@3$\uparrow$} & \makecell{Mean GT\\rank$\downarrow$} & Acc.$\uparrow$ & SR$\uparrow$ & \makecell{Valid\\JSON$\uparrow$} \\
\midrule
w/o Evol. & 96.0 & 62.2 & \textbf{60.0} & \textbf{0.72} & 29.05 & \textbf{97.4} & 79.5 & 87.2 \\
Evol. & \textbf{100.0} & \textbf{81.1} & 48.0 & 0.64 & \textbf{26.01} & \textbf{97.4} & \textbf{87.2} & \textbf{100.0} \\
\bottomrule
\end{tabular*}
\endgroup
\end{table}

%% file: sec/5_conclusion.tex
\vspace{-2.5mm}
\section{Limitations and Conclusion}
\paragraph{Limitations.}
Systematic comparisons of runtime, computational cost, and resource requirements against other agent frameworks and human experts remain open. Skill evolution is evaluated on a single benchmark with a limited number of tasks under a retrospective, in-domain setting; its effectiveness at larger scales and generalization to unseen targets and task families remain to be established. Experimental validation is also limited in scale and coverage of developability properties. Future work will expand evaluation across benchmarks and targets, conduct held-out assessments, and broaden experimental validation under matched conditions.

\paragraph{Conclusion.}
DrugHarness provides a mechanism for evolving scientific agent strategies through conditional skill learning. It turns execution experience into reusable guidance with explicit applicability conditions, independent scientific review, and feedback-driven revision. Evaluation on two benchmarks demonstrates broad molecular task capabilities, while retrospective results on skill-development tasks show improvements in screening, affinity comparison, and optimization. A prospective design study further links computational predictions to experimental activity, yielding a synthesized compound with improved cellular potency over all input references and activity competitive with selected expert-designed compounds. These findings support evidence-reviewed skill evolution as a promising approach to improving scientific agents and motivate broader evaluation of its generalization and experimental utility.

%% file: sec/x_supp.tex
\section{Benchmark and Evaluation Details}
\label{app:benchmark_details}
This section summarizes the task definitions and evaluation quantities of the source benchmarks. Table~\ref{tab:molbench} uses literature-reported baselines. The SMDD baselines combine published runs, with LO reaggregated on our fixed 40-task subset, and our Terra evaluation of the original harness.

\subsection{MolBench}
\label{app:molbench}
\textbf{Task coverage.} The 190 MolBench tasks comprise 35 property-filtering, 15 similarity-filtering, 37 affinity-comparison, 25 ranking, 39 editing, and 39 optimization tasks.

\textbf{Screening.}
MolBench-MS includes molecular property filtering, binding-affinity comparison, and docking-based screening \citep{zhang2026molclaw}. Property filtering selects molecules satisfying the supplied constraints from ten candidates. Binding-affinity comparison selects one of two molecules using a target-specific criterion, with experimental $K_i$ defining the reference order. Docking screening ranks 60 candidates per task, including experimentally active and inactive molecules. The source study defines mean Hits@3 as
\begin{equation}
\operatorname{MeanHits@3}=\frac{1}{n}\sum_{i=1}^{n}\left|\operatorname{Top3}(i)\cap A_i\right|,
\end{equation}
where $A_i$ is the active set for task $i$.

\textbf{Combined comparison.}
Table~\ref{tab:molbench} combines the Sonnet baselines reported in MolClaw with the Qwen baselines reported in CAi and the three DrugHarness NoSkill configurations. The first MolClaw-group column is the reported Claude Code baseline without MolClaw skills.

\textbf{Table notation.}
All proportion metrics in the MolBench tables are percentages. Acc., Prec., Spec., and Valid denote accuracy, precision, specificity, and validity; SR denotes success rate. Hit@3 is the percentage of ranking tasks with at least one active molecule in the top three, whereas Mean Hits@3 is the mean active-molecule count. Top-3 rate is Mean Hits@3 divided by three, expressed as a percentage; Count gives successful tasks over the total. Mean GT rank is the mean predicted rank of the reference top-three molecules. In affinity comparison, Valid measures valid candidate choices. In editing, Subst. denotes substitution and Overall aggregates editing operations. Optimization Valid is the valid-SMILES percentage; Hard and Soft denote the two scaffold-retention measures. Percentages are displayed to one decimal place; calculations retain the underlying values.

\textbf{Editing and optimization.}
MolBench-MO distinguishes functional-group editing from physicochemical property optimization. Editing accuracy is the fraction of outputs satisfying all requested modifications. Optimization concerns QED, LogP, or LogS and reports both a property-change quantity and success against predefined criteria. The editing experiment contains 39 tasks. Optimization success uses positive native improvement over 39 tasks. Table~\ref{tab:optimization_details} reports ordinary means of saved per-task gains separately by property.

\begin{table}[H]
\centering
\caption{\textbf{Property optimization with Codex/Terra.} Valid pairs denote valid scoring pairs; scaffold retention uses the hard criterion.}
\label{tab:optimization_details}
\small
\begin{tabularx}{\linewidth}{lYYYYY}
\toprule
Property & Tasks & Valid pairs & Positive gain & Mean gain & Scaffold retained \\
\midrule
LogP & 11 & 10 & 10 & 0.9260 & 4/11 \\
QED & 20 & 19 & 19 & 0.3748 & 3/20 \\
Solubility & 8 & 8 & 8 & 0.7089 & 7/8 \\
\bottomrule
\end{tabularx}
\end{table}
Across the 39 tasks, the hard scaffold-retention rate is 35.90\% and mean soft scaffold retention is 0.6360. These quantities describe structural preservation separately from the 94.9\% property-improvement success rate.

\subsection{Complete Evolution Results}
\label{app:evolution_details}
Table~\ref{tab:evolution_details} reports all 23 MolBench metrics for the Qwen3.7-Max configurations compared in Table~\ref{tab:evolution_summary}.
\input{tables/evolution_details}

\subsection{SMDD-Bench}
\label{app:smdd}
Our evaluation includes the complete Types 1--3 and 5 and the 40-task LO subset described below. Table abbreviations are 2D Pharm. (2D pharmacophore identification), IP (interaction point discovery), SH (scaffold hopping), LO (lead optimization), and FA (fragment assembly). The five source families contain 25, 25, 52, 340, and 60 instances, respectively \citep{han2026smdd}. The overall success rate is the fraction of successful instances across these families, rather than an unweighted average of family-level rates.

\textbf{2D pharmacophore identification.}
The input comprises a protein sequence and ten active and ten inactive molecules. The output is a Python predicate that classifies a supplied molecular SMILES string. Evaluation measures recall on hidden active molecules and specificity on hidden inactive molecules.

\textbf{Interaction point discovery.}
Given a receptor and pocket specification, the agent submits three typed three-dimensional points. Type-compatible points are matched one-to-one to reference interaction sites using a distance criterion of less than $2.5$\,\AA. Partial credit equals the number of matched points divided by three; success requires all three matches.

\textbf{Scaffold hopping.}
The output is a molecule with a different scaffold but preserved binding interactions. The detailed evaluator requires molecular validity, Morgan-fingerprint Tanimoto similarity below $0.5$, a scaffold maximum-common-substructure ratio below $0.65$, interaction-fingerprint similarity above $0.75$, and predicted binding probability above $0.7$.

\textbf{Lead optimization.}
The agent modifies a reference molecule while satisfying task-specific improvement targets, hold-constant intervals, and hard constraints. Evaluation jointly checks molecular validity, scaffold similarity, the property requirements, and predicted binding conditions. The reference-similarity threshold is at least $0.7$. Property directions and tolerances belong to individual task specifications.

\textbf{LO subset selection.}
DrugHarness uses a fixed subset of 40 LO tasks spanning 25 targets, with one or two tasks per target. Six previously selected tasks and a designated DHODH task were retained. Additional tasks were selected to cover optimization directions, objective counts, and structural input sizes using public task inputs, with natural task identifiers breaking ties. The subset contains 9/9/9/8/5 tasks with one/two/three/four/five objectives, respectively. Structure sizes comprise 10/18/12 tasks with at most 200, 201--500, and more than 500 ATOM-record residues.

\textbf{Fragment assembly.}
One or two spatially placed fragments must be retained within a valid output molecule. Evaluation checks fragment containment, connectivity where required, hard constraints, predicted binding probability above $0.7$, affinity within $0.1$ log units of the witness reference, and fragment RMSD below $2.0$\,\AA. Topological containment alone is insufficient.

\FloatBarrier
\clearpage
\section{Scientific Tools and Computational Environment}
\label{app:scientific_environment}

\subsection{Scientific Capabilities}
\label{app:scientific_capabilities}
DrugHarness provides a scientific environment spanning biomedical evidence retrieval, molecular design, and computational evaluation. Table~\ref{tab:scientific_capabilities} groups representative tools and resources by scientific function and summarizes their inputs and outputs. Operations can be invoked independently and combined according to the task, without imposing a fixed sequence of stages.

\input{tables/scientific_capabilities}

\clearpage
\section{Skill Learning Algorithms}
\label{app:learning_algorithms}
The main loop combines hypothesis induction, independent scientific review, version updates, and revision from subsequent-task feedback. The induction subroutine specifies how scientific conditions and evidence shape a proposed action change.

\begin{algorithm}[htbp]
\caption{Skill learning and revision from scientific experience}
\label{alg:evolution}
\begin{algorithmic}[1]
\Require Experiences $E$, insights $L$, skills $K$, history $V$, scope $\Omega$, budget $\mathcal B$
\Ensure Admitted skill versions, revised $L$, $E$, and $V$
\For{each development batch $B_t$ within $\Omega$ and $\mathcal B$}
 \State Incorporate trajectories and permitted native feedback already produced by task runs into $E$.
 \State $(\widetilde L,h)\gets\operatorname{Induce}(B_t,L,K,V;\Omega,\mathcal B_t)$ \Comment{One candidate branch}
 \State $L\gets\widetilde L$
 \If{$h=\varnothing$}
  \State Save $L$ and unresolved records; update $V$; \textbf{continue}.
 \EndIf
 \State $K'\gets\operatorname{Edit}(K,h)$; retain parent $K$, full content, sources, and intended change.
 \State An independent review agent checks sources, conditions, and executable instructions.
 \State Use the shared scientific tools or code for targeted local trials when needed; retain trial evidence.
 \State Complete $\operatorname{Review}(K',h,R_h;\Omega)$ and record $\rho$, or record incomplete status.
 \If{review completed and $\rho=\mathrm{admit}$}
  \State Record the admitted version $K'$ for subsequent tasks bound explicitly to that version.
  \State $K\gets K'$ for this branch.
 \Else
  \State Keep $K$; record rejection, deferral, or incomplete status.
 \EndIf
 \State Revise $L,V$ using the proposal and review records actually available.
 \If{the stage stopping condition is met}
  \State \textbf{break}
 \EndIf
\EndFor
\State Add subsequent task trajectories and permitted native feedback to $E$ when produced.
\end{algorithmic}
\end{algorithm}
Multiple proposal directions may be constructed and reviewed in parallel as separate branches. The update above describes one branch; its admission applies to the complete version reviewed, with no automatic merge of separately admitted candidates.

\begin{algorithm}[htbp]
\caption{Inducing a conditional action hypothesis}
\label{alg:induction}
\begin{algorithmic}[1]
\Require Batch $B_t$, insights $L$, skills $K$, history $V$, scope $\Omega$, allowance $\mathcal B_t$
\Ensure Revised insights $\widetilde L$ and one hypothesis $h$ or $\varnothing$
\State Identify recurring decision problems within $\Omega$ from actions and observations.
\State Separate contemporaneous observations, agent interpretations, and later feedback.
\State Check the scientific objects, endpoints, and result states.
\State Compare decision-relevant conditions; reread permitted sources for specific questions within $\mathcal B_t$.
\State Combine support for compatible conditions and consistent recommendations; retain separate scopes for different conditions.
\State For conflicts under comparable conditions, narrow or split guidance when justified; otherwise retain disagreement and uncertainty.
\State Compare with $K$ and $V$ to identify existing coverage and prior revision feedback.
\State Revise $L$ to obtain $\widetilde L$; propose $(g,c,a_0,a,R,U)$ for a testable gap, specifying the expected action difference.
\State Return one direction for this candidate branch and retain other directions for separate branches; return $\varnothing$ if no reusable modification is identified.
\end{algorithmic}
\end{algorithm}
\FloatBarrier

%% file: tables/evolution_details.tex
\begin{table}[H]
\centering
\caption{\textbf{Complete MolBench evolution metrics.} Both configurations use Codex/Qwen3.7-Max. Proportions are percentages; the mean $\Delta$ is the descriptive task-wise mean of saved native improvements.}
\label{tab:evolution_details}
\footnotesize
\begin{tabular}{lrr}
\toprule
Metric & w/o Evol. & Evol. \\
\midrule
\multicolumn{3}{c}{\textit{Screening}} \\
\midrule
Accuracy (\%)$\uparrow$ & 96.0 & 100.0 \\
Precision (\%)$\uparrow$ & 98.0 & 100.0 \\
Recall (\%)$\uparrow$ & 97.0 & 100.0 \\
F1 (\%)$\uparrow$ & 97.3 & 100.0 \\
Specificity (\%)$\uparrow$ & 100.0 & 100.0 \\
Validity (\%)$\uparrow$ & 98.0 & 100.0 \\
\midrule
\multicolumn{3}{c}{\textit{Affinity comparison}} \\
\midrule
Accuracy (\%)$\uparrow$ & 62.2 & 81.1 \\
Valid choice (\%)$\uparrow$ & 100.0 & 97.3 \\
\midrule
\multicolumn{3}{c}{\textit{Virtual screening}} \\
\midrule
Hit@3 (\%)$\uparrow$ & 60.0 & 48.0 \\
Mean Hits@3$\uparrow$ & 0.72 & 0.64 \\
Top-3 rate (\%)$\uparrow$ & 24.0 & 21.3 \\
Mean GT rank$\downarrow$ & 29.05 & 26.01 \\
Hit count$\uparrow$ & 15/25 & 12/25 \\
\midrule
\multicolumn{3}{c}{\textit{Editing}} \\
\midrule
Add (\%)$\uparrow$ & 100.0 & 100.0 \\
Delete (\%)$\uparrow$ & 100.0 & 100.0 \\
Substitute (\%)$\uparrow$ & 95.0 & 95.0 \\
Overall (\%)$\uparrow$ & 97.4 & 97.4 \\
Validity (\%)$\uparrow$ & 100.0 & 100.0 \\
\midrule
\multicolumn{3}{c}{\textit{Optimization}} \\
\midrule
Mean $\Delta$$\uparrow$ & 0.6054 & 0.7213 \\
Success rate (\%)$\uparrow$ & 79.5 & 87.2 \\
Valid SMILES (\%)$\uparrow$ & 87.2 & 100.0 \\
Scaffold hard (\%)$\uparrow$ & 38.5 & 48.7 \\
Scaffold soft (\%)$\uparrow$ & 61.8 & 69.0 \\
\bottomrule
\end{tabular}
\end{table}

%% file: tables/scientific_capabilities.tex
\begingroup
\small
\setlength{\tabcolsep}{5pt}
\renewcommand{\arraystretch}{1.08}
\setlength{\LTpre}{6pt}
\setlength{\LTpost}{6pt}
\begin{longtable}{@{}>{\raggedright\arraybackslash}p{0.19\linewidth}>{\raggedright\arraybackslash}p{0.405\linewidth}>{\raggedright\arraybackslash}p{\dimexpr0.405\linewidth-20pt\relax}@{}}
\caption{\textbf{Scientific capabilities of DrugHarness.} Representative tools and resources supporting over 80 agent-accessible scientific operations are organized into ten capability categories. A tool may support multiple operations; entries summarize capabilities rather than enumerate individual interfaces.}\label{tab:scientific_capabilities}\\
\toprule
\textbf{Capability} & \textbf{Tools and resources} & \textbf{Inputs $\rightarrow$ outputs} \\
\midrule
\endfirsthead
\multicolumn{3}{l}{\small\textit{Table~\ref{tab:scientific_capabilities} (continued)}}\\
\toprule
\textbf{Capability} & \textbf{Tools and resources} & \textbf{Inputs $\rightarrow$ outputs} \\
\midrule
\endhead
\midrule
\multicolumn{3}{r}{\small\textit{Continued on next page}}\\
\endfoot
\bottomrule
\endlastfoot
\textbf{Biomedical evidence retrieval} & Open Targets \citep{opentargets2025}; cBioPortal \citep{cbioportal2012,cbioportal2013}; TTD \citep{ttd2024}; ClinVar \citep{clinvar2025}; GTEx \citep{gtex2020}; ENCODE \citep{encode2020}; STRING \citep{string2023}; Europe PMC \citep{europepmc2024}; \href{https://clinicaltrials.gov/}{ClinicalTrials.gov} and \href{https://open.fda.gov/}{FDA} record interfaces. & Disease, target, variant, or experimental context $\rightarrow$ associations, variants, expression, functional and network evidence, publications, and study records with provenance. \\
\addlinespace[5pt]
\textbf{Target identity and structural data} & UniProt \citep{uniprot2025}; RCSB PDB \citep{rcsb2022}; SIFTS \citep{sifts2013}; AlphaFold DB \citep{alphafolddb2024}; NAKB (nucleic-acid structures) \citep{nakb2024}; KinHub \citep{kinhub_resource}. & Names, sequences, or database identifiers $\rightarrow$ target identities, sequences, structure candidates and coordinates, chain/residue mappings, and family annotations. \\
\addlinespace[5pt]
\textbf{Receptor, ligand, and pocket preparation} & PDBFixer \citep{pdbfixer_software}; Meeko \citep{meeko2025}; RDKit \citep{rdkit_software}; CCD \citep{ccd2015}; fpocket \citep{fpocket2009}; P2Rank \citep{p2rank2018}. & Protein/ligand structures, component selections, and preparation parameters $\rightarrow$ prepared receptors and ligand states, pocket candidates, docking grids, and input files. \\
\addlinespace[5pt]
\textbf{Interaction and pharmacophore analysis} & PLIP \citep{plip2025}; ProLIF \citep{prolif2021}; RDKit; atomic geometry analysis and co-crystal alignment/clustering. & Complexes, structural ensembles, or labeled molecules $\rightarrow$ contacts, water-mediated paths, interaction fingerprints, conserved three-dimensional interaction points, and molecular features. \\
\addlinespace[5pt]
\textbf{Compound sources and activity evidence} & ChEMBL \citep{chembl2024}; BindingDB \citep{bindingdb2025}; PubChem \citep{pubchem2025}; local-library similarity search; \href{https://patents.google.com/}{Google Patents} retrieval by publication number. & Target/compound identifiers, reference molecules, or patent publication numbers $\rightarrow$ chemical identities, measured activities with assay context, candidate rankings, or patent documents. \\
\addlinespace[5pt]
\textbf{Molecule generation and fragment linking} & DiffSBDD \citep{diffsbdd2024}; FlowMol3 \citep{flowmol3_2026}; AutoFragDiff \citep{autofragdiff2024}; explicit molecular-graph linking. & Generation settings, optional pocket context, or fragments with attachment sites $\rightarrow$ three-dimensional candidates, fragment mappings, and linked-molecule proposals. \\
\addlinespace[5pt]
\textbf{Docking, rescoring, and binding prediction} & AutoDock Vina \citep{vina2010,vina2021}; QuickVina2-GPU 2.1 \citep{vinagpu2024,vinagpu2023}; EquiScore \citep{equiscore2024}; Boltz-2 \citep{boltz2_2025,boltz1_2024}; MSA preparation \citep{colabfold2022}. & Receptors, ligands, grids/pockets, sequences, and required MSAs $\rightarrow$ poses, rescoring results, predicted complexes, binding-related values, and confidence estimates. \\
\addlinespace[5pt]
\textbf{Molecular processing and property assessment} & RDKit; ESOL (ChemCoTBench refit) \citep{esol2004,li2025chemcotbench}; PyTDC-compatible penalized LogP \citep{huang2021tdc,sascore2009}; ADMET-AI \citep{admetai2024}. & SMILES/SDF, candidate sets, and calculation parameters $\rightarrow$ standardized molecules, physicochemical properties, structural alerts, similarities, solubility, and ADMET predictions. \\
\addlinespace[5pt]
\textbf{Molecular editing and optimization proposals} & Explicit molecular-graph editing; REINVENT4 \citep{reinvent2024}; Mol2Mol \citep{mol2mol2022}; LibINVENT \citep{libinvent2022}. & Starting molecules, scaffolds, edit requests, or generation settings $\rightarrow$ edited candidates, scaffold decorations, analogues, and operation-integrity diagnostics. \\
\addlinespace[5pt]
\textbf{Retrosynthetic analysis} & AiZynthFinder \citep{aizynthfinder2024}. & Target molecules, stock, and search settings $\rightarrow$ candidate retrosynthetic routes, stock matches, and search results. \\
\end{longtable}
\endgroup

%% file: sec/x_case_studies.tex
\clearpage
\begingroup
\captionsetup{font=small,skip=5pt}
\section{Case Studies}
\label{app:case_studies}
\definecolor{baceblue}{HTML}{245A81}
\definecolor{baceorange}{HTML}{B77439}

These cases follow scientific work in DrugHarness from candidate assessment to guidance refinement: coordinating evaluations around a shared molecule, inspecting tool-generated structures, reusing computational feedback within a task, and revising guidance from prior execution and scientific review.

\subsection{Candidate-linked Assessment for BACE1 Optimization}
\label{app:bace1_joint}

Generating plausible molecules does not guarantee selecting a jointly feasible design. In SMDD-Bench, \citet{han2026smdd} report that retrospective evaluation of enumerated candidates recovers successes beyond the agents' original submissions.

Given a reference molecule and protein--ligand complex, the BACE1 task requires decreases of at least 0.1 from the supplied baselines in four predictions: blood--brain barrier penetration (BBB), cytochrome P450 3A4 (CYP3A4) inhibition, hERG potassium-channel blockade, and Ames mutagenicity (AMES). Candidates must also meet four property-holding conditions and physicochemical, structural and binding constraints. The task specifies these optimization directions and requires one candidate SMILES satisfying the joint requirements.

DrugHarness generated molecules with Python/RDKit, registered them as shared candidate objects, and obtained ADMET-AI predictions. The first two batches contained 22 structures including the reference. Figure~\ref{fig:bace1_tradeoffs} compares the reference with B2-09 and B3-02. Second-batch candidate B2-09 met three improvement targets but exceeded the CYP3A4 limit (Table~\ref{tab:bace1_properties}).

\begin{figure}[H]
\centering
\includegraphics[width=\linewidth]{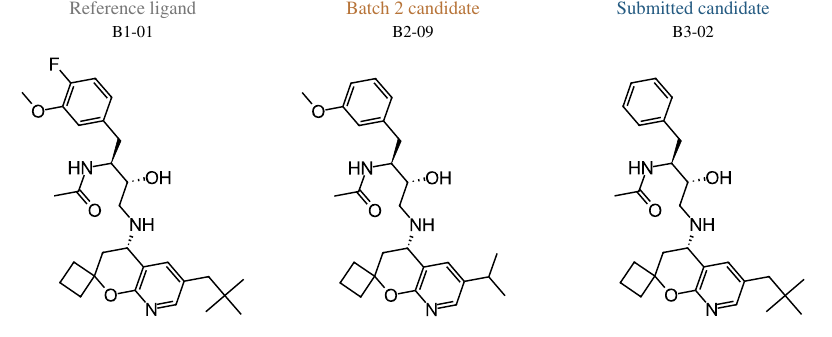}
\caption{\textbf{Reference and generated BACE1 candidates.} B2-09 and B3-02 were both generated from the reference molecule. Structures are aligned by their shared core.}
\label{fig:bace1_tradeoffs}
\end{figure}

\begin{table}[H]
\centering
\captionsetup{font=small,position=top,skip=7pt}
\captionsetup[subtable]{font=small,position=top,skip=4pt,justification=centering}
\caption{\textbf{ADMET-AI predictions for the BACE1 candidates.} B2-09 meets three of the four improvement targets but exceeds the CYP3A4 limit. B3-02 meets all four targets, and both candidates satisfy the four holding conditions.}
\label{tab:bace1_properties}
\input{tables/case_bace1_properties.tex}
\end{table}

\clearpage
\paragraph{Connecting assessment to the submitted object.}
After the second-batch predictions returned, DrugHarness generated ten further structures from the same reference. B3-02, the submitted candidate, met all four improvement targets and all four holding conditions. Its CYP3A4 prediction was 0.8079, compared with B2-09's 0.9311; its BBB, hERG and AMES predictions were higher but remained within their limits (Figure~\ref{fig:bace1_property_comparison}). Molecules from the generated SDF were registered as shared candidate records, allowing ADMET, structural comparison, and binding prediction to address the same molecule. For B3-02, RDKit computed a reference similarity of 0.7692, above the required 0.70; Boltz-2 returned a binding-probability prediction of 0.9944, above 0.70. DrugHarness then wrote and read back \texttt{solution.smi}, submitting the same molecular state used in the prediction (Figure~\ref{fig:bace1_objects}).

\begin{figure}[H]
\centering
\includegraphics[width=\linewidth]{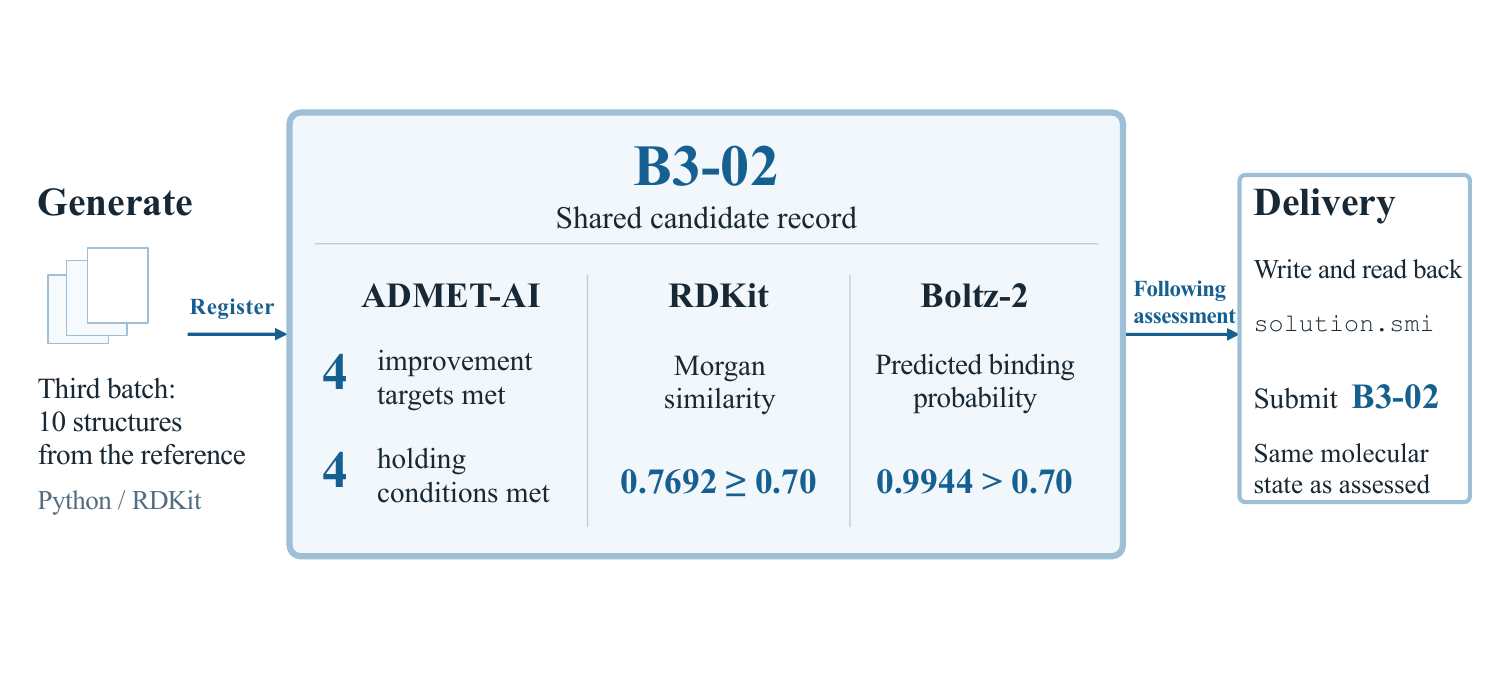}
\caption{\textbf{Candidate-linked assessment and submission.} The three assessments refer to the same registered B3-02 molecule. Following assessment, the candidate was written to \texttt{solution.smi}, read back and submitted in the evaluated molecular state.}
\label{fig:bace1_objects}
\end{figure}

The shared candidate representation links each assessment to the delivered molecular structure, allowing results from different scientific tools to be considered together under the task requirements.

\begin{figure}[H]
\centering
\includegraphics[width=\linewidth]{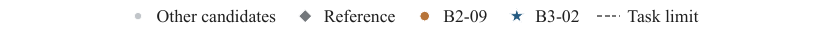}
\par\vspace{2pt}
\begin{subfigure}[t]{.49\linewidth}
\includegraphics[width=\linewidth]{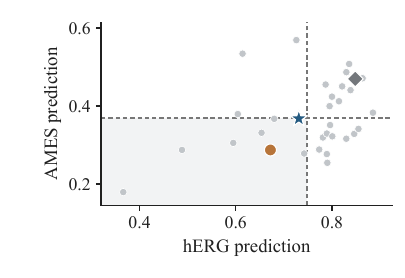}
\caption{Two-target view of the evaluated pool}
\end{subfigure}\hfill
\begin{subfigure}[t]{.49\linewidth}
\includegraphics[width=\linewidth]{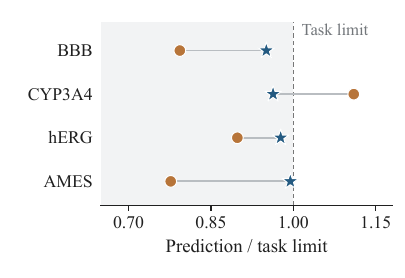}
\caption{Four-target trade-off for the highlighted pair}
\end{subfigure}
\caption{\textbf{Predicted property trade-offs among BACE1 candidates.} (a) All 32 evaluated structures, including the reference; shading marks the region meeting the hERG and AMES limits. (b) Four predictions divided by their task upper limits. Connected pairs compare B2-09 and B3-02 at each endpoint; ratios at or below one meet that target.}
\label{fig:bace1_property_comparison}
\end{figure}
\clearpage
\subsection{Structural Inspection during Scaffold Replacement}
\label{app:case_8okq}

Structural analysis can remain incomplete even when a predictor returns a complex. In the MolClaw run described by \citet{zhang2026molclaw_arxiv_v2}, ProLIF analysis failed during execution; the Met769 interaction was identified post hoc. In the 8OKQ task below, DrugHarness used the shared scientific workspace to inspect a returned complex before submission.

The SMDD-Bench 8OKQ scaffold-replacement task (Type~3) provides a reference ligand and a protein--ligand complex. It requires a new scaffold with Morgan similarity below 0.50, scaffold MCS ratio below 0.65, preservation of the reference binding interactions, and Boltz-2 binding probability above 0.70. The requested output is a candidate SMILES.

DrugHarness characterized ten candidate designs with local RDKit code. The selected cyclohexyl candidate had a Morgan similarity of 0.348 and a scaffold MCS ratio of 0.471 relative to the reference. Boltz-2 returned a binding-probability prediction of 0.9907 together with a complex structure. DrugHarness retrieved the predicted complex in CIF format and passed it to scientific code as the sole structure input. Figure~\ref{fig:8okq_handoff} traces this path from the candidate to the geometric analysis.

\begin{figure}[H]
\centering
\captionsetup[subfigure]{font=small,skip=3pt}
\begin{subfigure}[t]{\linewidth}
\centering
\begin{minipage}[t]{.48\linewidth}\centering
{\fontsize{9}{11}\selectfont\textcolor[HTML]{245A81}{Reference ligand}}\\[4pt]
\includegraphics[width=186bp,trim=0 10bp 0 10bp,clip]{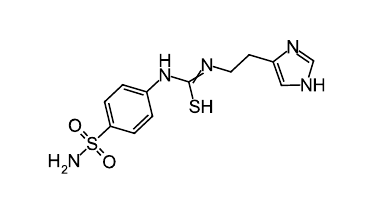}
\end{minipage}\hfill
\begin{minipage}[t]{.48\linewidth}\centering
{\fontsize{9}{11}\selectfont\textcolor[HTML]{B77439}{Submitted candidate}}\\[4pt]
\includegraphics[width=186bp,trim=0 10bp 0 10bp,clip]{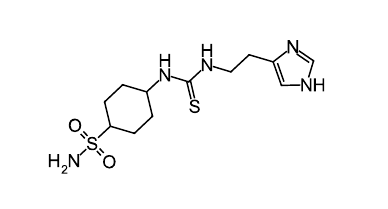}
\end{minipage}
\caption{Different scaffold, retained sulfonamide motif}
\end{subfigure}
\vspace{8pt}
\begin{subfigure}[t]{\linewidth}\centering
\includegraphics[width=\linewidth]{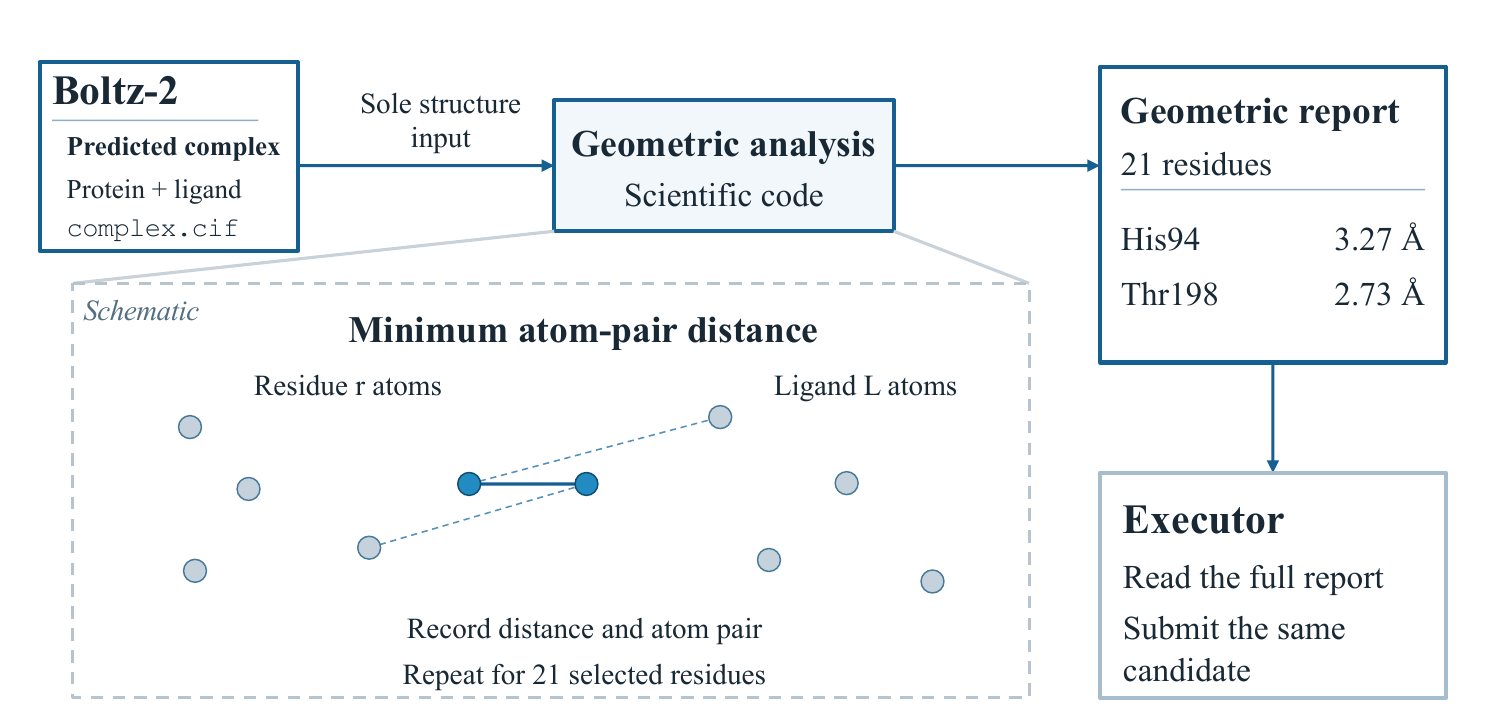}
\caption{A returned structure becomes an explicit input to scientific code}
\end{subfigure}
\caption{\textbf{Scaffold replacement and structural inspection.} (a) The reference ligand and submitted candidate have different ring scaffolds and share a sulfonamide group. (b) The predicted Boltz-2 complex was retrieved and used as the code's sole CIF input. For each selected residue $r$, the script computed the minimum distance to an atom of the predicted ligand $L$. The full 21-residue report was read before the same candidate was submitted. Arrows follow the recorded scientific handoffs.}
\label{fig:8okq_handoff}
\end{figure}

\clearpage
\paragraph{Geometric evidence returned before submission.}
The script calculated nearest-ligand distances for 21 selected protein residues and recorded the atom pair for each minimum. His94 NE2 and Thr198 OG1 lay 3.27 and 2.73~\AA{} from ligand atom N32, respectively. DrugHarness read the complete report before writing and submitting the unchanged candidate. Figure~\ref{fig:8okq_geometry} places the recorded distances alongside the predicted pocket geometry.

\begin{figure}[H]
\centering
\captionsetup[subfigure]{font=small,skip=3pt}
\begin{subfigure}[t]{.49\linewidth}\centering
\vspace{0pt}
\begin{minipage}[c][1.35436\linewidth][c]{\linewidth}
\includegraphics[width=\linewidth]{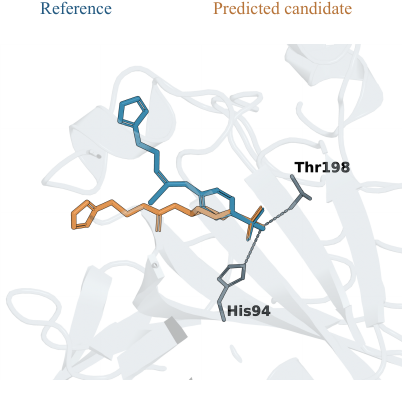}
\end{minipage}
\caption{Pocket overlay for spatial context}
\end{subfigure}\hfill
\begin{subfigure}[t]{.49\linewidth}\centering
\vspace{0pt}
\begin{minipage}[c][1.35436\linewidth][c]{\linewidth}
\includegraphics[width=\linewidth]{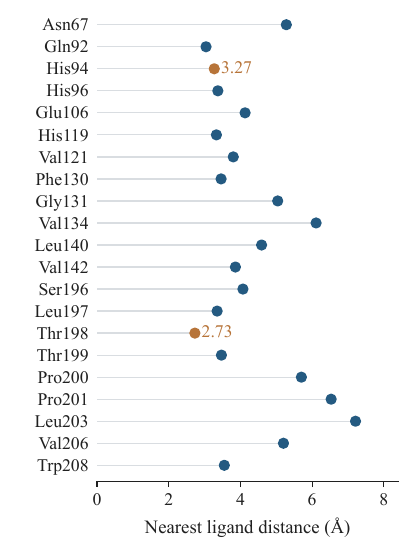}
\end{minipage}
\caption{Distances for all 21 selected residues}
\end{subfigure}
\caption{\textbf{Local static geometry of the predicted complex.} Panel (b) shows the nearest-ligand distances recorded during task execution. The pocket overlay in (a) uses structures aligned after execution for visualization. Orange points identify His94 and Thr198; the remaining selected residues are blue.}
\label{fig:8okq_geometry}
\end{figure}

DrugHarness carried the predicted complex from a specialized tool into scientific code and returned the geometric report to the executor before submission. Links between the candidate record, predicted complex, code input and derived report keep this sequence traceable to the submitted molecule.
\clearpage
\subsection{QED Optimization through Iterative Feedback}
\label{app:case_qed}
Given a source SMILES, this MolBench task asks for an increase in the quantitative estimate of drug-likeness (QED) through intermediate optimization steps. QED improvement is the evaluation criterion, with no target-activity preservation constraint; the final SMILES is submitted in JSON.

Starting from a molecule with recorded QED 0.4982, DrugHarness generated and evaluated three candidate batches using Python/RDKit within the same task. The first batch yielded a leading candidate with QED 0.9285. The next script used that exact evaluated structure as its baseline to construct further variants. The second batch yielded an N-methyl-amide candidate with QED 0.9375, which the third script likewise used as its baseline before exploring additional substitutions.

DrugHarness selected a third-batch candidate with QED 0.9394, verified its molecular representation, and submitted the same structure in the requested JSON answer. The archived tables contain 83 records: 80 numerical QED entries and three invalid molecular strings, with repeated candidates retained. Figure~\ref{fig:qed_case} shows the recorded scores and the evaluated structures reused as starting points for subsequent optimization scripts.

\begin{figure}[H]
\centering
\captionsetup[subfigure]{font=small,skip=3pt}
\begin{subfigure}[t]{\linewidth}
\centering
\includegraphics[width=\linewidth]{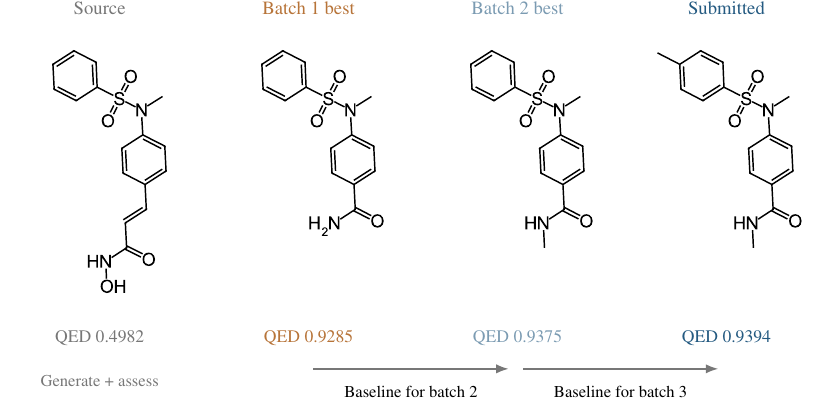}
\caption{Evaluated molecules reused as design baselines}
\end{subfigure}

\vspace{4pt}
\begin{subfigure}[t]{\linewidth}
\centering
\includegraphics[width=\linewidth]{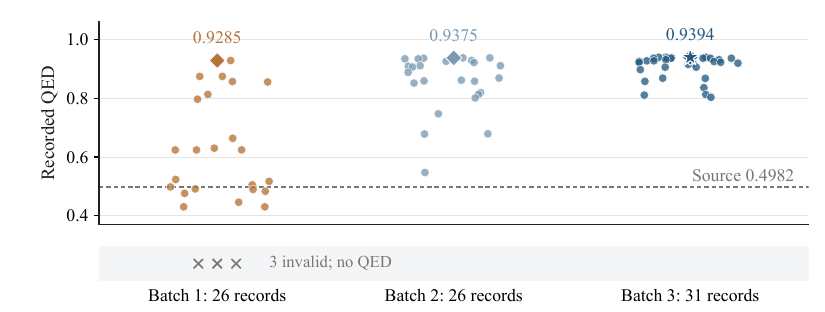}
\caption{Recorded candidate scores across three batches}
\end{subfigure}

\caption{\textbf{QED feedback reused during molecular optimization.} (a) The source, first- and second-batch leaders, and submitted candidate, drawn from the recorded molecular structures. Arrows indicate reuse of evaluated structures as baselines in subsequent scripts. (b) All 80 numerical entries from batches containing 26, 26, and 31 records, including repeated structures. Orange, light blue, and dark blue indicate batches 1--3. Diamonds mark batch maxima, the star marks the submitted candidate, and the dashed line denotes the source QED. Three invalid first-batch entries occupy a separate non-numerical strip. The submitted candidate shares the highest third-batch QED.}
\label{fig:qed_case}
\end{figure}
\clearpage
\subsection{Scientific Review for Affinity Skill Refinement}
\label{app:case_affinity}

The MolBench affinity comparison labeled 012 here specifies CXCR2 as the target and provides two candidate SMILES. It asks for the weaker binder (higher $K_i$), with the selected molecule returned as a SMILES; evaluation checks the candidate choice. Alongside this scientific task, we compare how executors using the initial skill (S0) and its revised version (S1) gathered evidence and chose a candidate.

S0 prioritized exact activity records. When candidate mappings were unavailable, it suggested trying another query or source if budget permitted. During review of S0, an independent review agent scanned complete BindingDB sources and matched both candidates in each of the source tasks labeled 003 and 021, recovering the measurements in Table~\ref{tab:affinity_review_measurements}. Subsequent executions still showed incomplete source recovery or candidate mapping. Researchers selected these execution records and the review's trial results for a focused learning round.

\begin{table}[H]
\centering
\captionsetup{font=small,position=top,skip=7pt}
\caption{\textbf{Affinity records recovered by the first independent review.} The reviewer matched both candidate structures in each source task; A/B labels are task-specific.}
\label{tab:affinity_review_measurements}
\input{tables/case_affinity_review_measurements.tex}
\end{table}

The learner explicitly cited the independent checks and automatically authored S1. When result previews lack candidate mappings, S1 specifies full-source recovery, matching both query structures and extracting their measured activities. A second independent review applied this procedure to 003 and 021 and recovered both candidates' activity records in each task (Figure~\ref{fig:affinity_learning}).

\begin{figure}[H]
\centering
\includegraphics[width=\linewidth]{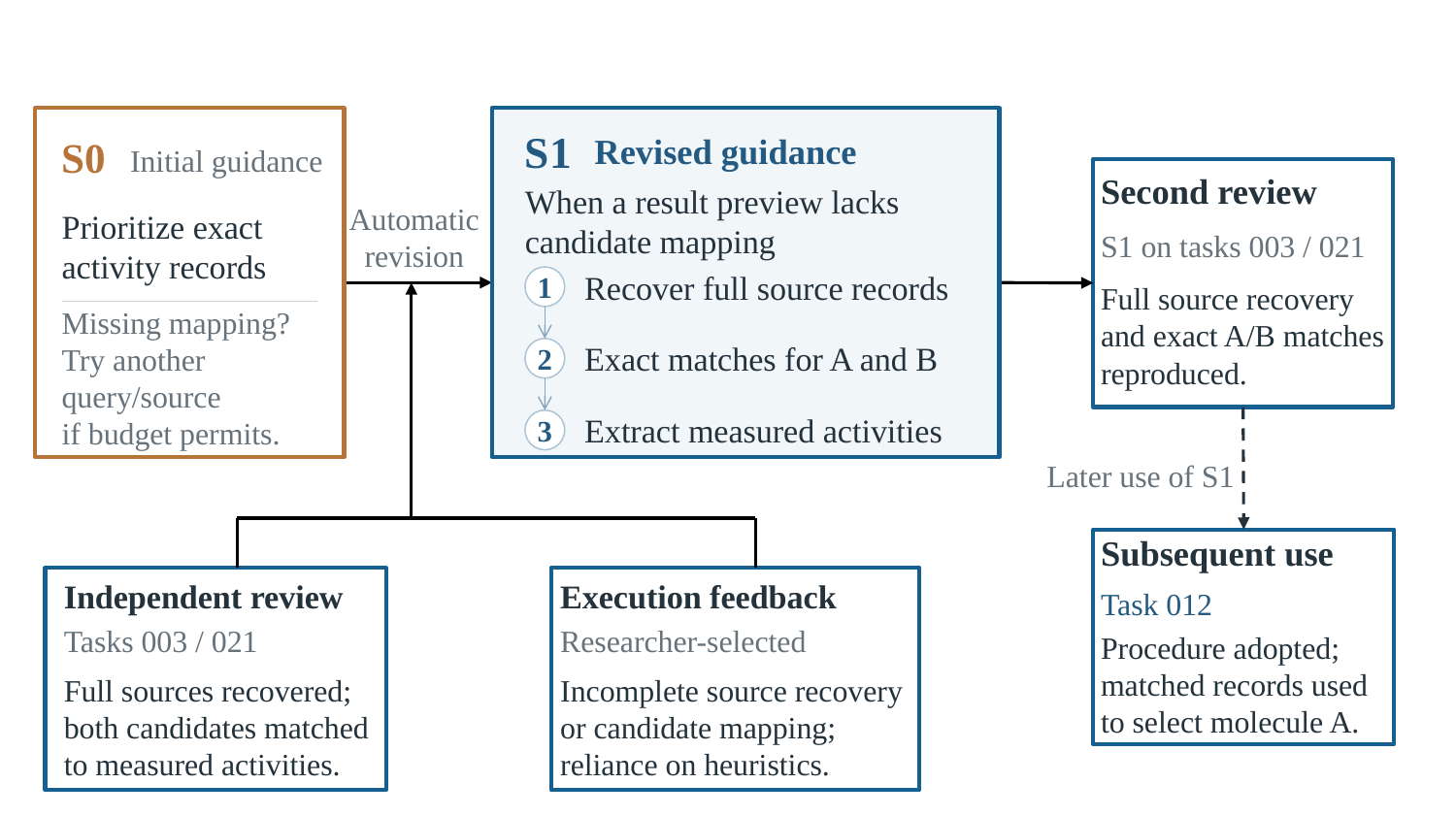}
\caption{\textbf{From scientific feedback to revised affinity guidance.} Independent review results and researcher-selected execution feedback informed the S0-to-S1 revision. The added recovery-and-matching procedure was exercised in a second review on 003 and 021 before its use on task 012.}
\label{fig:affinity_learning}
\end{figure}

\clearpage
\paragraph{Revised guidance adopted in a subsequent task.}
Task 012 was outside this focused learning input. The executions using S0 and S1 both received sources containing the relevant measurements. In the S1 execution, the executor read the revised skill and its reference checklist, then followed the source-recovery and structure-matching procedure to obtain candidate-linked activity records.

Using these records, the S1 executor selected molecule A, the weaker binder with the higher $K_i$ (Figure~\ref{fig:affinity_adoption}). The S0 executor selected B after partial source reads and reliance on property or conformer heuristics. Table~\ref{tab:affinity_execution} summarizes the evidence use and resulting choices.

\begin{figure}[H]
\centering
\includegraphics[width=\linewidth]{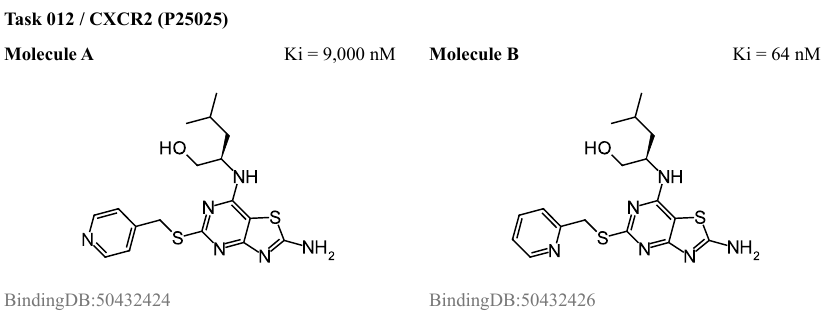}
\caption{\textbf{CXCR2 candidates and matched affinity records.} Molecular structures are task inputs; $K_i$ values were retrieved from BindingDB records associated with \citet{karlstrom2013cx3cr1}.}
\label{fig:affinity_adoption}
\end{figure}

\begin{table}[H]
\centering
\captionsetup{font=small,position=top,skip=7pt}
\caption{\textbf{Evidence use and candidate selection in task 012.} The initial skill (S0) and revised skill (S1) were each used in one execution of the CXCR2 affinity task, which asks for the weaker binder (higher $K_i$).}
\label{tab:affinity_execution}
\input{tables/case_affinity_execution_comparison.tex}
\end{table}
\endgroup

%% file: tables/case_bace1_properties.tex
\begingroup
\fontsize{9}{11}\selectfont
\renewcommand{\arraystretch}{1.12}
\setlength{\tabcolsep}{0pt}
\newcolumntype{R}[1]{>{\raggedleft\arraybackslash}p{#1}}
\begin{subtable}[t]{0.48\linewidth}
\centering
\caption{Improvement targets}
\label{tab:bace1_improvement}
\begin{tabular*}{\linewidth}{@{\extracolsep{\fill}}p{48pt}R{50pt}R{42pt}R{42pt}@{}}
\toprule[0.7pt]
Endpoint & Task limit & \textcolor{baceorange}{B2-09} & \textcolor{baceblue}{B3-02}\\
\midrule[0.4pt]
BBB & $\leq 0.7680$ & 0.6093 & 0.7303\\
CYP3A4 & $\leq 0.8387$ & \textcolor{baceorange}{\bfseries 0.9311} & \textcolor{baceblue}{\bfseries 0.8079}\\
hERG & $\leq 0.7482$ & 0.6720 & 0.7310\\
AMES & $\leq 0.3696$ & 0.2871 & 0.3676\\
\addlinespace[3pt]
Met & & $3/4$ & $4/4$\\
\bottomrule[0.7pt]
\end{tabular*}
\end{subtable}\hfill
\begin{subtable}[t]{0.48\linewidth}
\centering
\caption{Holding conditions}
\label{tab:bace1_holding}
\begin{tabular*}{\linewidth}{@{\extracolsep{\fill}}p{48pt}R{50pt}R{42pt}R{42pt}@{}}
\toprule[0.7pt]
Endpoint & Task limit & \textcolor{baceorange}{B2-09} & \textcolor{baceblue}{B3-02}\\
\midrule[0.4pt]
Solubility & $\geq -4.8248$ & $-4.1327$ & $-4.5324$\\
Clearance & $\leq 47.5618$ & 39.0512 & 43.8456\\
Caco-2 & $\geq -5.3597$ & $-5.1867$ & $-5.1951$\\
PPB & $\leq 95.2947$ & 89.5430 & 94.9833\\
\addlinespace[3pt]
Met & & $4/4$ & $4/4$\\
\bottomrule[0.7pt]
\end{tabular*}
\end{subtable}
\endgroup

%% file: tables/case_affinity_review_measurements.tex
\begingroup
\fontsize{9}{11}\selectfont
\renewcommand{\arraystretch}{1.15}
\setlength{\tabcolsep}{14pt}
\begin{tabular}{@{}lrrr@{}}
\toprule[0.7pt]
Source task & Records scanned & $K_i$(A), nM & $K_i$(B), nM\\
\midrule[0.4pt]
003 & 2,008 & 7.9 & 2,512\\
021 & 195 & 9.5 & 2,300\\
\bottomrule[0.7pt]
\end{tabular}
\endgroup

%% file: tables/case_affinity_execution_comparison.tex
\begingroup
\fontsize{9}{11}\selectfont
\renewcommand{\arraystretch}{1.15}
\begin{tabular*}{\linewidth}{@{\extracolsep{\fill}}>{\raggedright\arraybackslash}p{.21\linewidth}>{\raggedright\arraybackslash}p{.30\linewidth}>{\raggedright\arraybackslash}p{.41\linewidth}@{}}
\toprule[0.7pt]
Comparison & \textcolor{baceorange}{Initial skill (S0)} & \textcolor{baceblue}{Revised skill (S1)}\\
\midrule[0.4pt]
Source processing & Partial source reads & Full activity file recovered; 1,523 records scanned\\
\addlinespace[6pt]
Decision evidence & Property or conformer heuristics & One matched record per molecule; corresponding $K_i$ values\\
\addlinespace[6pt]
Selected molecule & B (incorrect) & A (correct)\\
\bottomrule[0.7pt]
\end{tabular*}
\endgroup